%% file: iclr2027_conference.tex
\documentclass{article}
\usepackage{iclr2027_conference,times}
\input{math_commands.tex}

\usepackage{hyperref}
\usepackage{url}
\usepackage{graphicx}
\usepackage{colortbl}
\definecolor{methodgray}{gray}{0.93}
\usepackage{booktabs}
\usepackage{longtable}

\title{DeSpatial: The devil of Agentic Spatial Reasoning Lies in 3D Itself}
\author{%
{\small\textbf{Yifei Liu\textsuperscript{1,2}\thanks{Equal contribution. \quad \textsuperscript{\textdagger}Corresponding author.}\hspace{0.4em} Xiaolong Zhou\textsuperscript{1}\footnotemark[1]\hspace{0.4em} Ziyang Gong\textsuperscript{3}\hspace{0.4em} Yifan Zhou\textsuperscript{3}\hspace{0.4em} Yuanyuan Gao\textsuperscript{4}}}\\
{\small\textbf{Yuning Gong\textsuperscript{5}\hspace{0.4em} Yuchen Yang\textsuperscript{6}\hspace{0.4em} Hongjie Zhang\textsuperscript{1}\hspace{0.4em} Xue Yang\textsuperscript{3}\hspace{0.4em} Xiao Sun\textsuperscript{1}\hspace{0.4em} Zhihang Zhong\textsuperscript{3}\textsuperscript{\textdagger}}}\\[2pt]
{\normalfont\small\textsuperscript{1}Shanghai Artificial Intelligence Laboratory \quad \textsuperscript{2}Beihang University}\\
{\normalfont\small\textsuperscript{3}Shanghai Jiaotong University \quad \textsuperscript{4}HKUST \quad \textsuperscript{5}Tencent \quad \textsuperscript{6}Fudan University}
}
\iclrfinalcopy

\begin{document}
\maketitle
\lhead{Preprint}
\begin{abstract}
\input{sections/abstract}
\end{abstract}
\input{sections/introduction}
\input{sections/diagnostic}
\input{sections/method}
\input{sections/experiments}

\input{sections/related_work}
\input{sections/conclusion}
\clearpage
\bibliography{iclr2027_conference}
\bibliographystyle{iclr2027_conference}
\clearpage
\appendix
\input{sections/appendix}
\end{document}

%% file: math_commands.tex
\usepackage{amsmath,amsfonts,bm}

\def\eqref#1{equation~\ref{#1}}

\def\1{\bm{1}}

\DeclareMathAlphabet{\mathsfit}{\encodingdefault}{\sfdefault}{m}{sl}
\SetMathAlphabet{\mathsfit}{bold}{\encodingdefault}{\sfdefault}{bx}{n}

%% file: sections/abstract.tex
Spatial agents have access to reconstruction and segmentation tools, yet inconsistent predictions and fragmented object identities can prevent their outputs from becoming reliable 3D scene information. A simple oracle experiment exposes this perception bottleneck: replacing tool-predicted perception with ground-truth 3D bounding boxes raises an agent to nearly full score across five object-centric VSI-Bench tasks, while retaining the same reasoning tools. Motivated by this finding, we believe the devil of spatial reasoning lies in 3D itself, and introduce DeSpatial(\textbf{De}vil of \textbf{Spatial}), which explicitly decomposes spatial understanding into perception and reasoning. DeSpatial has a shared 3D perception tool to reconcile independent perception outputs into a unified, metric, gravity-aligned 3D scene. This perception tool has two core components: a) a \textit{joint segmentation--depth co-refinement} module that consolidates masks of the same object and removes projection artifacts caused by inconsistent segmentation and depth within a single frame; b) a \textit{cross-view instance fusion} module that consolidate voxel-overlapping instances but not allowed to merge same-frame instances. This fusion module additionally fuses spatially disjoint fragments caused by insufficient multi-view observations, by checking their 3D structure and source mask appearance. After receiving the complete and unified 3D scene information, the agent chooses reasoning or computational tools to solve the spatial problem. DeSpatial achieves leading performance on VSI-Bench, MindCube-Tiny, and ViewSpatial, with additional long-video results on VSI-SUPER, revealing the bottleneck of spatial understanding lies in a precise 3D perception itself.

%% file: sections/introduction.tex
\section{Introduction}
\label{sec:introduction}

Spatial understanding requires a consistent account of a scene: how many objects it contains, how large they are, and where they lie relative to one another or to an observer. Videos make this problem particularly demanding because the same object can appear repeatedly, with different visible surfaces and intervening occluders. Recent benchmarks have exposed these challenges through various types of visual questions including object measurements, multi-view questions, and changes of perspective~\citep{yang2025thinking,wang2025mindcube,li2025viewspatial}.

Spatial agents address these demands by combining vision-language models (VLMs) with perception tools, such as 3D reconstruction~\citep{wang2025vggt}, segmentation~\citep{kirillov2023sam}, and detection~\citep{liu2025groundingdino}. Access to these tools, however, does not automatically provide reliable 3D scene understanding. Their independently predicted masks, depths, and reconstructed surfaces need to be reconciled into objects with consistent geometry and identity. Existing approaches organize spatial evidence through visual programs~\citep{yuan2024visual,luo2026pyspatial}, rendered views~\citep{zhang2026think3d}, or structured scene memories~\citep{mou2026visra,he2026reasmory}. Yet even a correct sequence of tool calls can leave an agent with conflicting evidence. For example, inconsistencies between a chair's 2D mask and independently predicted depth can introduce floor points into the chair's 3D point cloud during back-projection, as illustrated in Figure~\ref{fig:joint_corefinement}(b). Observations of different parts of the same table may also be treated as separate instances, as illustrated in Figure~\ref{fig:instance_fusion}(b). The resulting errors propagate into otherwise valid counting, measurement, and relational computations. We therefore conjecture that effective agentic spatial understanding depends on both how to organize intermediate perception outputs to a 3D representation and how accurately this representation describes the scene.

To understand this problem better, we do a simple oracle experiment that reveals how much the quality of this 3D representation matters. We use ground-truth (GT) annotated 3D bounding boxes as the outputs of a pseudo 3D perception tool and pair the agent with suitable spatial reasoning and computational tools to solve tasks. On VSI-Bench, the agent can achieve a surprising score of 98.23\% across five object-centric tasks covering 4,030 questions, as shown in Table~\ref{tab:oracle}, while the performance drops greatly when replacing the oracle 3D tool with real perception tools. The comparison identifies perception as a major bottleneck in these tasks and motivates improving the 3D evidence available to the reasoner. Section~\ref{sec:diagnostic} details the annotation interface and scope of this diagnostic.

Motivated by this observation, we introduce \textbf{DeSpatial}, a framework that explicitly decomposes spatial understanding into perception and reasoning. It shares the same 3D perception tool across all spatial tasks, taking responsibility for reconciling independent 2D perception and 3D reconstruction outputs. This tool returns a unified, metric, gravity-aligned scene representation containing object identities, oriented bounding boxes, and source observations. A separate set of reasoning and computation tools can be dynamically chosen to operate on this shared representation to answer the question. This division gives the agent a consistent set of objects and spatial attributes to work with, reducing the need to reconcile intermediate perception outputs during task reasoning.

Our perception design addresses two coupled problems. \emph{Joint segmentation--depth co-refinement} first consolidates duplicate masks and identifies parts of an occluded object within each frame. It then combines semantic membership with geometric reliability to remove erroneous projections by using a sequence of geometric and component filters, as shown in Figure~\ref{fig:joint_corefinement}. \emph{Cross-view instance fusion} establishes identity across frames. Voxel overlap provides positive association evidence, while distinct instances observed in the same frame prohibit incompatible merges. For cross-view observations of different object parts that fully disjoint in space, a visual judge combines multiple source views with their 3D arrangement to decide whether they belong to one physical object, as shown in Figure~\ref{fig:instance_fusion}. These operations consolidate observed evidence into a scene that downstream tools can consistently measure and query.

Our contributions are threefold:
\begin{itemize}
    \item We identify the accuracy of 3D perception as a major bottleneck in agentic spatial understanding through an oracle experiment: given GT 3D boxes, an agent achieves strong performance with the same reasoning and computational tools. This locates the ``devil'' in the accuracy of the underlying 3D information: improving perception can unlock the agent's existing reasoning capability without a more elaborate dynamic tool-calling mechanism.
    \item We develop a shared 3D perception tool that addresses inconsistencies between independently predicted segmentation and depth. Its cross-view fusion uses voxel-overlap matching to recover associations between partial observations, while same-frame exclusions guard against merging distinct objects, jointly targeting high recall and precision. Visual identity verification further identifies fragments of the same object from spatially non-overlapping observations.
    \item DeSpatial achieves leading agentic spatial-understanding performance on VSI-Bench, MindCube-Tiny, and ViewSpatial, and extends the evaluation to long-video spatial understanding on VSI-SUPER.
\end{itemize}

%% file: sections/diagnostic.tex
\begin{figure}[!t]
\centering
\includegraphics[width=\linewidth]{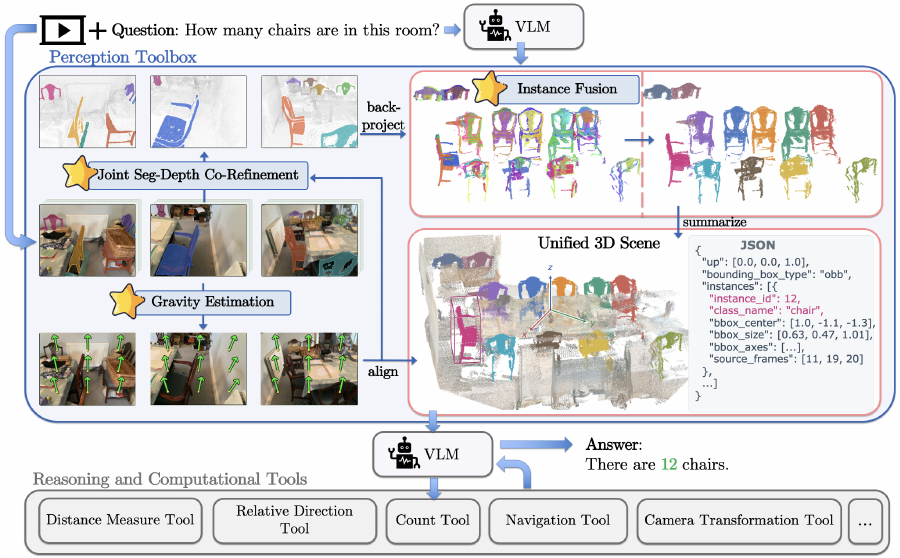}
\caption{\textbf{Overview of DeSpatial.} The VLM receives the video and question and calls our shared perception tool. Joint segmentation--depth co-refinement cleans per-frame observations; cross-view instance fusion associates them into distinguished individual objects. Colors identify individual masks before fusion and persistent instances afterward. Gravity estimation establishes a common vertical direction. The VLM then reasons and answers using reasoning and computational tools.}
\label{fig:method_overview}
\end{figure}

\section{Diagnosing the Perception Bottleneck}
\label{sec:diagnostic}

We test how well a spatial agent can use reasoning tools when the object geometry needed for a question is supplied by ground truth annotations. At the perception--reasoning interface, we supply an oracle 3D tool that returns ground-truth 3D oriented bounding boxes (OBBs). These boxes are the only 3D information provided by the oracle tool, and the agent still requests object classes, selects instances, calls spatial computation tools, and derives the answer through ordinary tool calls by itself. The provided reasoning and computational tools are the same as in Section~\ref{sec:reasoning}.

\paragraph{Evaluation setup.}
We compare the oracle 3D tool with a raw-perception baseline that constructs object point clouds by directly back-projecting pixels within SAM3~\citep{carion2025sam3} masks using Pi3~\citep{wang2025pi3} geometry. Both conditions use Qwen3-VL-8B-Instruct~\citep{bai2025qwen3vl}, the same system prompt, tool registrations, iteration limits, and 32-frame sampling policy, and all reasoning and computational tools are kept the same.
Both the oracle and raw-perception conditions evaluate the same 4,030 questions across five VSI-Bench categories: object counting, absolute distance, object size, relative distance, and relative direction. Numerical questions use mean relative accuracy (MRA), and multiple-choice questions use accuracy. Appendix~\ref{sec:oracle_protocol} details the experiment settings.

\input{tables/oracle_diagnostic}

\paragraph{What the oracle reveals.}
Table~\ref{tab:oracle} shows that with the oracle 3D tool, scores on all five tasks are close to perfect. This result should not be over-explained because the annotations in VSI-Bench are not perfect~\citep{zhang2026revsi}, and the annotations align with the benchmark's questions' answers. However, it suffices to indicate that given the proper 3D information, the agent's spatial reasoning ability is little constrained by how to utilize the spatial measurement, analysis, or relative direction tools, but rather identifies perception as a major bottleneck. It motivates improving the scene representation consumed by those tools, rather than  simply providing the agent with a large list of perception and reasoning tools and expecting it to magically know how to fully reconcile the inconsistency between various perception results. Next, we introduce our framework focusing on improving the 3D perception ability inspired by this oracle experiment.

%% file: tables/oracle_diagnostic.tex
\begin{table}[t]
\centering
\begingroup
\small
\setlength{\tabcolsep}{5pt}
\renewcommand{\arraystretch}{1.12}
\caption{Oracle and raw-perception results on the same 4,030 questions across five VSI-Bench categories.}
\label{tab:oracle}
\begin{tabular}{@{}l*{6}{r}@{}}
\toprule
3D Tool & \shortstack{Obj.\\Count} & \shortstack{Abs.\\Dist.}
& \shortstack{Obj.\\Size} & \shortstack{Rel.\\Dist.}
& \shortstack{Rel.\\Dir.} & Avg. \\
\midrule
SAM3 + Pi3 (raw) & 42.70 & 32.54 & 49.68 & 47.62 & 79.37 & 50.38 \\
GT 3D Boxes & 98.76 & 96.73 & 99.37 & 98.59 & 97.70 & \textbf{98.23} \\
\bottomrule
\end{tabular}
\endgroup
\end{table}

%% file: sections/method.tex
\section{DeSpatial}
\label{sec:method}

DeSpatial separates spatial understanding into perception and reasoning (Figure~\ref{fig:method_overview}). Its perception tool constructs a unified, gravity-aligned 3D scene through geometry estimation, joint segmentation--depth co-refinement, and cross-view instance fusion; reasoning tools then operate on this representation.

\subsection{Metric and Gravity-Aligned Geometry}
\label{sec:metric_geometry}
We obtain dense geometry and camera poses with reconstruction models~\citep{wang2025vggt,wang2025pi3,lin2025depthanything3}, and rescale them to approximate metric scale using Depth Anything 3~\citep{lin2025depthanything3}.
We uniformly sample up to 12 frames and estimate their camera-coordinate up directions with GeoCalib~\citep{veicht2024geocalib}. Let $R_t$ denote the camera-to-world rotation of frame $t$, and let $\hat{\mathbf{u}}_t^{c}$ be its predicted unit up vector, opposite to gravity. We express these estimates in the common reconstruction coordinates as
\begin{equation}
\hat{\mathbf{u}}_t^{w}=R_t\hat{\mathbf{u}}_t^{c}.
\label{eq:gravity_world}
\end{equation}
To reject inconsistent estimates, we select the direction with the most neighbors within an angular threshold $\theta=20^\circ$:
\begin{equation}
\mathcal{I}(j)=\left\{t:\arccos\!\left((\hat{\mathbf{u}}_t^{w})^{\top}\hat{\mathbf{u}}_j^{w}\right)\leq\theta\right\},
\qquad j^{\star}=\arg\max_j|\mathcal{I}(j)|.
\label{eq:gravity_consensus}
\end{equation}
Ties are resolved by the smaller median inlier angle. We use $\mathcal{I}=\mathcal{I}(j^{\star})$ for joint calibration; if fewer than $\min(4,N)$ frames agree, where $N$ is the number sampled, we fall back to all sampled frames and flag the estimate as less reliable.

GeoCalib then jointly calibrates the selected views with shared intrinsics and fixed camera rotations. Their refined camera-coordinate directions $\tilde{\mathbf{u}}_t^{c}$ are constrained by a single unit scene-up vector $\mathbf{u}_{\mathrm{raw}}$:
\begin{equation}
\tilde{\mathbf{u}}_t^{c}=R_t^{\top}\mathbf{u}_{\mathrm{raw}},
\qquad \|\mathbf{u}_{\mathrm{raw}}\|_2=1,
\qquad t\in\mathcal{I}.
\label{eq:gravity_joint}
\end{equation}
We recover the shared direction by normalizing $\sum_{t\in\mathcal{I}}R_t\tilde{\mathbf{u}}_t^{c}$. Finally, we construct a rotation $R$ such that $R\mathbf{u}_{\mathrm{raw}}=\mathbf{e}_z$, and apply the same metric scale $s$ and rotation to points and camera centers, $\mathbf{p}'=sR\mathbf{p}$ and $\mathbf{c}_t'=sR\mathbf{c}_t$; camera rotations become $R_t'=RR_t$. This places the scene in metric coordinates with $+z$ pointing upward, providing a common frame for reconciling the independently predicted masks and depths.

\subsection{Joint Segmentation--Depth Co-Refinement}
\label{sec:co_refinement}
\begin{figure}[t]
\centering
\includegraphics[width=\linewidth]{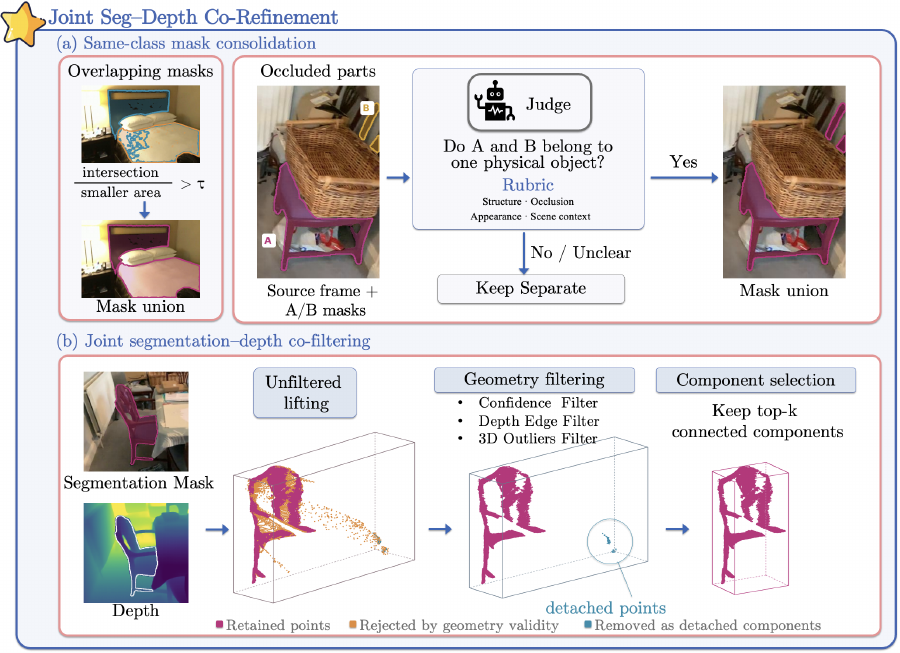}
\caption{\textbf{Joint segmentation--depth co-refinement.} (a) Same-class mask consolidation unifies highly overlapping masks and visually verified parts of an occluded object. Negative or uncertain identity judgments keep masks separate. (b) Confidence, depth-edge, and 3D outlier filters remove unreliable geometry. Component selection then removes residual detached points that distort the bounding box, retaining the top-$k$ components for $k$ confirmed object parts, usually one.}
\label{fig:joint_corefinement}
\end{figure}

The module addresses inconsistencies within a single frame. It first consolidates same-class masks within each frame, then filters their geometric support (Figure~\ref{fig:joint_corefinement}). Consolidation addresses two sources of repeated object identities. For highly overlapping masks, we take their union when the intersection exceeds half the smaller mask's area. Occlusion can instead split one object into disjoint visible parts. We use their gravity-aligned horizontal voxel projections to identify overlapping candidates, then ask a VLM judge whether the marked parts in the source image belong to one physical object. The judge considers visible structure, the occluder, appearance, and scene context. Only a positive decision permits mask union. Resolving these identities before cross-view fusion is essential: otherwise, the same-frame exclusion constraint in the fusing stage would incorrectly keep parts of one object apart.

A consolidated mask can still include incorrect 3D points. Independent segmentation and depth predictions often disagree around object boundaries, assigning background geometry to foreground masks, as shown in the chair back region in Figure~\ref{fig:joint_corefinement}(b). Even a small contaminated region can inflate the estimated object extent and distort size or distance measurements. We therefore combine semantic membership with reconstruction confidence, depth-discontinuity checks, and 3D outlier filtering. For consolidated mask $M_{t,i}$, the geometry-filtered point set is
\begin{equation}
P_{t,i}=\left\{sR\mathbf{p}_t(\mathbf{x})\;\middle|\;
\mathbf{x}\in M_{t,i}\cap V_t\right\},
\label{eq:filtered_proposal}
\end{equation}
where $\mathbf{p}_t$ is the reconstructed world-coordinate point map, $sR$ applies metric scaling and gravity alignment, and $V_t$ is the set of pixels passing the geometric filters detailed in Appendix~\ref{sec:geometry_filtering}.

These filters remove many outliers but cannot reliably reject an entire coherent background patch admitted by a mask. In Figure~\ref{fig:joint_corefinement}(b), disagreement around the openings in a chair back leaves detached floor points even after geometry filtering. We therefore select connected components within each proposal. A mask comprising $k$ visible parts, confirmed in the mask consolidation phase, retains its $k$ largest 3D components; ordinary masks use $k=1$. Duplicate-mask union alone does not increase $k$. Selecting $k$ components allows multiple confirmed parts to remain while removing smaller detached contamination.

\subsection{Cross-View Instance Fusion}
\label{sec:cross_view_fusion}
Clean per-frame observations still need persistent global object identities. One way is to use spatial clustering such as DBSCAN~\citep{ester1996density,ruan2026allospatial} or centroid-distance clustering~\citep{mou2026visra}, but is prone to merging nearby objects. We also tried to merge based on 3D bounding-box IoU, but find it is sensitive to partial visibility and may wrongly connect objects whose bounding boxes overlap but geometry don't. We instead associate occupied voxels while preserving evidence that two observations represent different objects.

Each refined observation becomes a node carrying its class, source frame $t_i$, and point set. Let $A_i$ and $A_j$ be the occupied voxel sets of two same-class observations from different frames. We measure their overlap by
\begin{equation}
w_{ij}=\frac{|A_i\cap A_j|}{\min(|A_i|,|A_j|)}.
\label{eq:voxel_overlap}
\end{equation}
Normalizing by the smaller occupied set accommodates observations that cover different portions of an object. We retain candidate edges with $w_{ij}\geq\tau_{\mathrm{ov}}$ and process them in descending score order.

\begin{figure}[t]
\centering
\includegraphics[width=\linewidth]{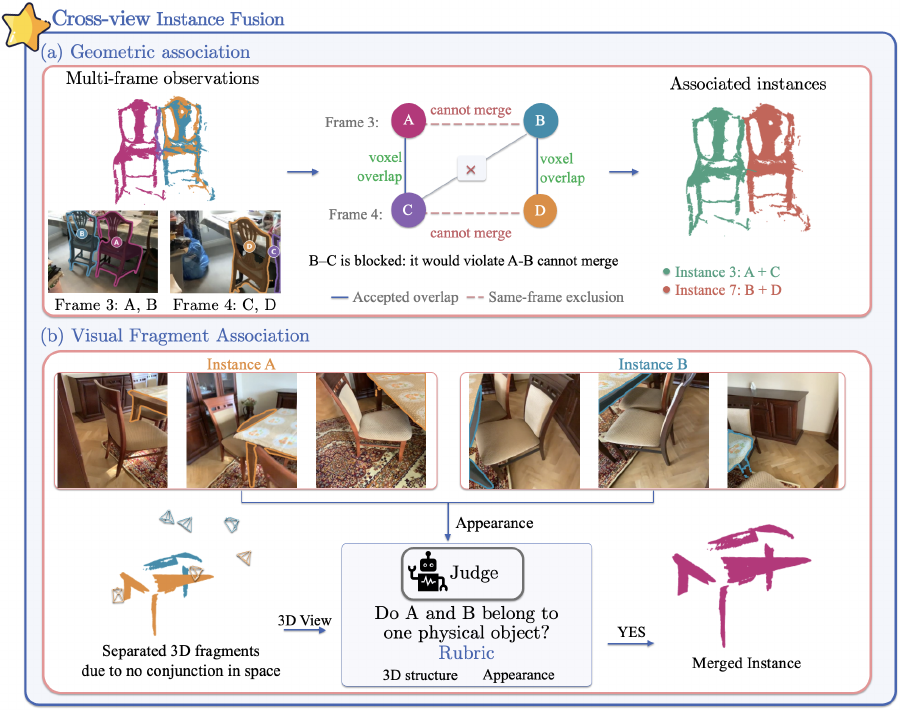}
\caption{\textbf{Cross-view instance fusion.} (a) Voxel overlap links partial observations, while same-frame exclusions prevent distinct objects from merging, including through indirect connections. (b) Visual fragment association handles observations without shared voxels. Multiple source views and a common 3D view provide complementary appearance and structural evidence to an identity judge. A positive decision combines the observed fragments and their provenance.}
\label{fig:instance_fusion}
\end{figure}

To prevent distinct objects from merging, we impose \emph{same-frame exclusion} on entire groups of observations. For a group $G$, let $\mathcal{F}(G)=\{t_i:i\in G\}$ be its set of source frames. An edge can join groups $G_a$ and $G_b$ only when
\begin{equation}
\mathcal{F}(G_a)\cap\mathcal{F}(G_b)=\emptyset.
\label{eq:same_frame_exclusion}
\end{equation}
Thus, distinct consolidated observations from one frame cannot enter the same instance, including through indirect chains of overlapping observations. In Figure~\ref{fig:instance_fusion}(a), joining A with C blocks a later B--C merge because A and B were observed as distinct objects. Appendix~\ref{sec:association_details} gives the voxel size, overlap threshold, and implementation details.

\label{sec:fragments}
However, voxel overlap alone cannot merge all observations of an object: different viewpoints may capture only spatially disjoint parts, as shown in Figure~\ref{fig:instance_fusion}(b). We therefore retrieve same-class fragment pairs from the union of neighbor sets defined by 3D surface distance and gravity-aligned horizontal projection distance, and use a VLM to judge whether they belong to the same physical object. The VLM receives several masked source views of each candidate together with their 3D visualization in the common coordinate frame, and assesses structural compatibility and appearance to decide whether to merge them. Details of the VLM judge are provided in Appendix~\ref{sec:association_details}.

The final scene packages these fused instances in one consistent representation. Each instance has a persistent identifier, semantic class, centroid, gravity-aligned oriented bounding box (center, side lengths, and axes), and source frame observations. Together with the scene up direction and camera geometry, these attributes let downstream tools refer to the same objects without independently reconciling masks, depths, or cross-view identities.

\subsection{Reasoning with the Unified Scene}
\label{sec:reasoning}

The VLM uses structured tool calls to count objects, measure dimensions and distances, determine relative directions, transform reference frames, plan routes, and identify appearance order from the shared scene representation and source observations. Appendix~\ref{sec:reasoning_details} describes these operations and their reference-frame conventions.

%% file: sections/experiments.tex
\section{Experiments}
\label{sec:experiments}

\subsection{Evaluation Setup}
Qwen3-VL-8B-Instruct serves as the agent backbone. We evaluate video-based spatial understanding on VSI-Bench, multi-view mental modeling on MindCube-Tiny, and perspective-dependent reasoning on ViewSpatial~\citep{yang2025thinking,wang2025mindcube,li2025viewspatial}. VSI-SUPER further tests spatial recall and continual counting over long videos~\citep{yang2025cambrian}.

\subsection{Main Results}
DeSpatial achieves an average score of 69.22 on VSI-Bench (Table~\ref{tab:vsi-bench}), demonstrating that its shared scene representation supports diverse spatial computations. It obtains 87.19 overall accuracy on MindCube-Tiny and 68.35 on ViewSpatial (Table~\ref{tab:mindcube-viewspatial}), extending its performance to multi-view reasoning and changes of perspective. On VSI-SUPER (Table~\ref{tab:vsi-super}), DeSpatial maintains high recall across increasing video durations and improves continual counting at longer horizons. These benchmarks extend the evaluation beyond object measurements to changes of viewpoint and observations accumulated over long videos. Together, these results show the utility of consistent scene information across object-centric, viewpoint-dependent, and long-video spatial reasoning.

\input{tables/vsi_bench_results}
\input{tables/mindcube_viewspatial_results}
\input{tables/vsi_super_results}

\subsection{Ablation Study}
\label{sec:ablations}
\label{sec:backbones}
\label{sec:failure_analysis}

We evaluate joint segmentation--depth co-refinement and cross-view instance fusion on a fixed 504-question subset of VSI-Bench, and compare gravity estimators on 156 scenes. Joint segmentation--depth co-refinement improves the average score by 11.27 points over raw mask lifting (Table~\ref{tab:corefinement_ablation}(a)). Geometric filtering suppresses inconsistent geometric observations and improves the average score by 5.34 points, while largest-component selection removes residual detached geometry and yields a further 2.32-point gain. Adding same-class mask consolidation and part-aware top-$k$ selection brings another 3.61-point improvement, demonstrating the benefit of consolidating and preserving valid parts of the same object within each frame (Table~\ref{tab:corefinement_ablation}(a)). For cross-view instance fusion, using voxel overlap for instance association achieves a higher average score than using AABB IoU or DBSCAN clustering (Table~\ref{tab:instance_fusion_ablation}(b)). Same-frame exclusion improves the average score by 2.52 points by preventing erroneous merges between distinct objects. Visual fragment association yields a further 2.01-point gain by linking spatially disjoint observations of the same object (Table~\ref{tab:instance_fusion_ablation}(b)). For gravity estimation, we visualize GeoCalib joint calibration and a RANSAC~\mbox{\citep{fischler1981random}} plane-based estimator on the same 156 scenes for human assessment (Table~\ref{tab:gravity_ablation}). GeoCalib achieves 100\% human-assessed correctness on the 156 evaluated scenes, compared with 68.59\% for the RANSAC estimator. The human annotation setup and interface examples are provided in Appendix~\ref{sec:gravity_human_evaluation}.

\input{tables/corefinement_ablation}
\input{tables/gravity_ablation}

%% file: tables/vsi_bench_results.tex
\begin{table}[t]
\centering
\begingroup
\footnotesize
\setlength{\tabcolsep}{1.95pt}
\renewcommand{\arraystretch}{1.12}
\caption{VSI-Bench results (\%): MRA for numerical questions and accuracy for multiple choice. Avg. averages eight categories (four multiple-choice categories for Think3D). Unmarked methods use the full benchmark. $^{*}$Subset evaluations from the original papers.}\label{tab:vsi-bench}
\begin{tabular}{@{}ll*{9}{r}@{}}
\toprule
& & \multicolumn{4}{c}{Numerical (MRA)}
& \multicolumn{4}{c}{Multiple choice (accuracy)} & \\
\cmidrule(lr){3-6}\cmidrule(lr){7-10}
Method & Backbone
& \shortstack{Obj.\\Count} & \shortstack{Abs.\\Dist.}
& \shortstack{Obj.\\Size} & \shortstack{Room\\Size}
& \shortstack{Rel.\\Dist.} & \shortstack{Rel.\\Dir.}
& \shortstack{Route\\Plan} & \shortstack{Appr.\\Order} & Avg. \\
\midrule
\addlinespace[3pt]
\multicolumn{11}{l}{\textit{Trained models}} \\*
Cambrian-S & Qwen2.5-7B & 73.2 & 50.5 & 74.9 & 72.2 & 71.1 & 76.2 & 41.8 & 80.1 & 67.5 \\
VLM-3R & LLaVA-NV-7B & 70.2 & 49.4 & 69.2 & 67.1 & 65.4 & 80.5 & 45.4 & 40.1 & 60.9 \\
CaMo-3B & Qwen2.5-VL-3B & 64.8 & 18.6 & 59.2 & 43.2 & 45.2 & 45.3 & 25.8 & 49.9 & 44.0 \\
\addlinespace[3pt]
\multicolumn{11}{l}{\textit{Training-free inference pipelines~\citep{li2025seetrek,ma2026reason}}} \\*
See\&Trek & InternVL3-8B & 65.2 & 32.9 & 46.9 & 46.7 & 45.9 & 40.2 & 30.4 & 37.4 & 43.2 \\
ReRe & InternVL3-8B & 38.8 & 31.1 & 44.4 & 30.0 & 37.2 & 37.3 & 30.9 & 23.6 & 34.16 \\
\addlinespace[3pt]
\multicolumn{11}{l}{\textit{Spatial reasoning agents}} \\*
ViSRA & InternVL3-8B
& 36.6 & 34.6 & 58.8 & 52.7 & 47.3 & 62.4 & 46.9 & 59.7 & 49.88 \\
AlloSpatial$^{*}$ & Gemini-3-Pro
& 51.8 & 36.8 & 57.7 & 62.6 & 62.0 & 67.7 & 65.3 & 83.7 & 61.0 \\
Think3D$^{*}$ & Gemini-2.5-Pro & -- & -- & -- & -- & 54.00 & 37.30 & 46.93 & 68.24 & 51.61 \\
\midrule
\rowcolor{methodgray}
\textbf{DeSpatial} & Qwen3-VL-8B
& 65.43 & 55.08 & 66.86 & 73.58 & 67.93 & 82.53 & 64.43 & 77.90 & 69.22 \\
\bottomrule
\end{tabular}
\par
\endgroup
\end{table}

%% file: tables/mindcube_viewspatial_results.tex
\begin{table}[t]
\centering
\setlength{\abovecaptionskip}{2pt}
\setlength{\belowcaptionskip}{0pt}
\begingroup
\newcommand{\benchmethod}[2]{\mbox{#1\raisebox{-0.5ex}{\scalebox{0.65}{\mdseries #2}}}}
\fontsize{8}{9.6}\selectfont
\setlength{\tabcolsep}{0.9pt}
\renewcommand{\arraystretch}{1.12}
\caption{Results on MindCube-Tiny and ViewSpatial. Scores are accuracy percentages; Avg. denotes overall accuracy. Model references are listed in Appendix~\ref{sec:baseline_models_table3}.}\label{tab:mindcube-viewspatial}\label{tab:mindcube}\label{tab:viewspatial}
\begin{minipage}[t]{0.49\linewidth}
\centering
\fontsize{7.5}{9}\selectfont
\textbf{(a) MindCube}\par
\resizebox{\linewidth}{!}{%
\begin{tabular}{@{}l*{4}{r}@{}}
\toprule
Method & Among & Around & Rotation & Avg. \\
\midrule
Qwen2.5-VL-3B & 36.00 & 45.20 & 34.00 & 37.81 \\
GPT-4o & 43.00 & 46.40 & 35.00 & 42.29 \\
Gemini-2.5-Pro & 43.80 & 67.20 & 88.50 & 57.90 \\
Gemini-3-Pro & 68.20 & 77.20 & 93.00 & 75.10 \\
\benchmethod{VADAR}{GPT-4o} & 40.67 & 46.80 & 33.50 & 40.76 \\
\benchmethod{pySpatial}{GPT-4o} & 66.33 & 71.20 & 41.00 & 62.67 \\
\benchmethod{GCA}{Qwen3-VL-Thinking} & 59.80 & 61.80 & 82.00 & 64.20 \\
\benchmethod{AlloSpatial}{Gemini-3-Pro} & 75.80 & 86.00 & 93.50 & 81.60 \\
\benchmethod{Think3D}{Gemini-2.5-Pro} & 49.67 & 62.75 & 87.00 & 59.97 \\
\midrule
\rowcolor{methodgray}
\benchmethod{\textbf{DeSpatial}}{Qwen3-VL-8B} & 85.08 & 86.80 & 94.00 & 87.19 \\
\bottomrule
\end{tabular}%
}
\end{minipage}\hfill
\begin{minipage}[t]{0.49\linewidth}
\centering
\fontsize{7.5}{9}\selectfont
\textbf{(b) ViewSpatial}\par
\resizebox{\linewidth}{!}{%
\begin{tabular}{@{}l*{6}{r}@{}}
\toprule
& \multicolumn{2}{c}{Camera} & \multicolumn{3}{c}{Person} & \\
\cmidrule(lr){2-3}\cmidrule(lr){4-6}
Method & \shortstack{Rel.\\Dir.} & \shortstack{Obj.\\Ori.} & \shortstack{Rel.\\Dir.} & \shortstack{Obj.\\Ori.} & \shortstack{Scene\\Sim.} & Avg. \\
\midrule
Gemini-2.5-Pro & 54.30 & 24.30 & 47.27 & 59.84 & 39.28 & 46.09 \\
Qwen3-VL-8B & 54.20 & 29.70 & 40.30 & 47.30 & 31.10 & 42.20 \\
Qwen2.5-VL-7B & 46.64 & 29.72 & 35.04 & 37.05 & 28.78 & 36.85 \\
InternVL2.5-8B & 49.41 & 41.27 & 42.04 & 46.79 & 32.85 & 43.24 \\
\benchmethod{SenseNova-SI}{InternVL3-8B} & 66.30 & 43.20 & 43.10 & 38.40 & 70.00 & 54.70 \\
\benchmethod{S-Agent}{GPT-5.4} & 62.50 & 55.50 & 81.10 & 42.20 & 60.60 & 60.00 \\
\benchmethod{MV-STRIDE}{Qwen3-VL-8B} & 62.55 & 33.13 & 39.43 & 48.09 & 46.33 & 48.35 \\
\benchmethod{CaMo}{Qwen2.5-VL-3B} & 47.50 & 30.90 & 31.50 & 53.90 & 35.10 & 41.00 \\
\benchmethod{SpatialLadder}{Qwen2.5-VL-3B} & 48.30 & 24.10 & 34.40 & 71.10 & 38.90 & 44.20 \\
\midrule
\rowcolor{methodgray}
\benchmethod{\textbf{DeSpatial}}{Qwen3-VL-8B} & 65.71 & 81.22 & 69.00 & 78.01 & 51.76 & 68.35 \\
\bottomrule
\end{tabular}%
}
\end{minipage}
\endgroup
\end{table}

%% file: tables/vsi_super_results.tex
\begin{table}[t]
\centering
\setlength{\abovecaptionskip}{2pt}
\setlength{\belowcaptionskip}{0pt}
\begingroup
\footnotesize
\setlength{\tabcolsep}{3pt}
\renewcommand{\arraystretch}{1.12}
\caption{Results on VSI-SUPER. Recall (VSR) uses accuracy and Count (VSC) uses mean relative accuracy (MRA), both in percent; durations are in minutes. Model references and result sources are listed in Appendix~\ref{sec:baseline_models_table4}.}
\label{tab:vsi-super}
\begin{tabular}{@{}l*{9}{r}@{}}
\toprule
& \multicolumn{5}{c}{Recall (VSR): accuracy}
& \multicolumn{4}{c}{Count (VSC): MRA} \\
\cmidrule(lr){2-6}\cmidrule(lr){7-10}
Method & 10 & 30 & 60 & 120 & 240 & 10 & 30 & 60 & 120 \\
\midrule
MovieChat & 18.3 & 21.7 & 16.7 & 26.7 & 25.6 & 0.0 & 0.0 & 0.0 & 0.0 \\
Flash-VStream & 28.3 & 33.3 & 23.3 & 28.3 & 31.7 & 0.0 & 0.0 & 0.0 & 0.0 \\
Cambrian-S-7B & 38.3 & 35.0 & 6.0 & 0.0 & 0.0 & 0.0 & 0.0 & 0.0 & 0.0 \\
Cambrian-S-7B-LFP & 45.0 & 41.7 & 40.0 & 40.0 & 40.0 & 40.6 & 42.0 & 35.0 & 34.0 \\
S-Agent & 75.0 & 55.0 & 63.3 & 66.1 & 77.2 & 10.6 & 4.2 & 0.0 & 0.0 \\
\midrule
\rowcolor{methodgray}
\textbf{DeSpatial} & 90.3 & 88.3 & 90.1 & 88.3 & 86.7 & 33.4 & 53.4 & 49.4 & 47.2 \\
\bottomrule
\end{tabular}
\par
\endgroup
\end{table}

%% file: tables/corefinement_ablation.tex
\begin{table}[htbp]
\centering
\setlength{\abovecaptionskip}{2pt}
\setlength{\belowcaptionskip}{0pt}
\caption{Ablations on a fixed stratified VSI-Bench subset (504 questions, 63 per category); Avg. averages eight categories. (a) Full cross-view fusion is fixed. (b) Full co-refinement is fixed; Exclusion: same-frame exclusion; FragAssoc.: visual fragment association. All runs share the agent settings. IoU/voxel thresholds: 0.3/0.1; voxel size: 0.1\,m.}
\label{tab:corefinement_ablation}
\label{tab:instance_fusion_ablation}
\begin{minipage}[t]{0.52\linewidth}
\centering
\footnotesize
\textbf{(a) Joint seg--depth co-refinement}\par
\setlength{\tabcolsep}{3pt}
\begin{tabular}{@{}lr@{}}
\toprule
Configuration & Avg (\%) \\
\midrule
Raw mask lifting & 57.56 \\
\quad + Geometry filtering & 62.90 \\
\quad + Largest component ($k=1$) & 65.22 \\
\quad + Same-class mask consolidation & 66.45 \\
\quad + Part-aware top-$k$ selection & 68.83 \\
\bottomrule
\end{tabular}
\end{minipage}\hfill
\begin{minipage}[t]{0.46\linewidth}
\centering
\footnotesize
\textbf{(b) Cross-view instance fusion}\par
\setlength{\tabcolsep}{2pt}
\begin{tabular}{@{}lccr@{}}
\toprule
Overlap & Exclusion & FragAssoc. & Avg (\%) \\
\midrule
AABB IoU & Yes & No & 59.90 \\
DBSCAN & Yes & No & 61.37 \\
Voxel overlap & No & No & 64.30 \\
Voxel overlap & Yes & No & 66.82 \\
Voxel overlap & Yes & Yes & 68.83 \\
\bottomrule
\end{tabular}
\end{minipage}
\end{table}

%% file: tables/gravity_ablation.tex
\begin{table}[!ht]
\centering
\setlength{\abovecaptionskip}{2pt}
\setlength{\belowcaptionskip}{0pt}
\small
\setlength{\tabcolsep}{10pt}
\caption{Human assessment of projected up directions on the same 156 scenes.}
\label{tab:gravity_ablation}
\begin{tabular}{@{}lrr@{}}
\toprule
Gravity estimator & Correct / Total & Correctness (\%) \\
\midrule
RANSAC plane-based & 107 / 156 & 68.59 \\
GeoCalib joint calibration & 156 / 156 & \textbf{100.00} \\
\bottomrule
\end{tabular}
\end{table}

%% file: sections/related_work.tex
\section{Related Work}
\label{sec:related_work}
Spatial understanding with VLMs requires recovering object properties, spatial relationships, and changes of viewpoint from visual observations. VSI-Bench evaluates these abilities through video-based spatial questions, MindCube examines mental modeling from limited views, and ViewSpatial studies reasoning across camera and human perspectives~\citep{yang2025thinking,wang2025mindcube,li2025viewspatial}. Existing non-agentic approaches improve either the learned visual representations or the spatial evidence available during inference. Cambrian-S strengthens video representations for spatial understanding, while VLM-3R aligns reconstructive representations with language instructions to incorporate geometric information into VLMs~\citep{yang2025cambrian,fan2026vlm3r}. Complementary approaches introduce spatial cues without updating model parameters. See\&Trek supplies spatial prompting cues, and ReRe revisits initial answers using synthesized complementary views~\citep{li2025seetrek,ma2026reason}. These methods demonstrate different ways to expose spatial structure that may be difficult to infer from the original observations alone. Their emphasis ranges from learning geometry-aware representations to improving the evidence presented at inference time. Agentic approaches extend this direction by making perception and spatial computation available as executable operations, enabling models to gather and process evidence over multiple steps.

Agentic spatial reasoning augments VLMs with specialist perception and geometric computation tools. Beyond selecting an operation, an agent must compose these tools and interpret intermediate outputs such as images, point clouds, camera poses, and object geometry. Existing methods approach this process through program generation, interactive exploration, geometric constraints, and reusable skills. VADAR synthesizes a dynamic Python API, while pySpatial generates programs for scene reconstruction, camera manipulation, and novel-view rendering~\citep{marsili2025visual,luo2026pyspatial}. Think3D supports interactive exploration of reconstructed 3D space, allowing the agent to gather visual evidence from additional viewpoints~\citep{zhang2026think3d}. GCA formalizes reference frames and geometric objectives to guide tool selection and computation~\citep{chen2025gca}. Skill-3D distills scene-aware skills from execution trajectories, whereas NeSy-Spatial evolves reusable tool-use and geometric skills~\citep{li2026skill3d,tian2026nesyspatial}. SpatialCLI learns tool-use policies and internalizes capabilities acquired through tool interaction~\citep{zhou2026spatialcli}.

Other agents emphasize organizing spatial evidence for downstream reasoning. MSSR curates a minimal sufficient set through interacting perception and reasoning agents~\citep{guo2025pursuing}, while AlloSpatial constructs allocentric landmark and route maps from reconstructions filtered by confidence and point density~\citep{ruan2026allospatial}. These approaches highlight the importance of making perception outputs usable by an agent. DeSpatial focuses on the accuracy and consistency of the underlying 3D information: independent segmentation and reconstruction outputs can disagree geometrically, and observations of one object can remain fragmented within or across views. Our perception tool resolves these issues before returning a unified scene representation.

%% file: sections/conclusion.tex
\section{Conclusion}

DeSpatial treats reliable 3D scene construction as a central requirement for spatial agents. Its oracle study shows that an agent with the same reasoning tools can solve object-centric spatial questions substantially more accurately when given GT 3D bounding boxes. This motivates a shared perception tool that resolves inconsistencies between segmentation and geometry, consolidates object observations within and across frames, and returns unified scene information for subsequent computation. Joint segmentation--depth co-refinement and cross-view instance fusion address geometric contamination and fragmented identity together, thus achieving strong results on multiple spatial understanding benchmarks.

\paragraph{Limitations.}
The representation depends on detections, reconstructed geometry, metric scale, and gravity estimates. Component selection cannot recover unobserved surfaces or missing detections, and incorrect part counts or visual identity judgments can discard valid geometry or merge different objects. The oracle results concern five object-centric categories and answer-consistent annotations; they do not establish that perception is the only limitation across all spatial tasks.

%% file: sections/appendix.tex
\section{Oracle Protocol}
\label{sec:oracle_protocol}

\paragraph{Information supplied.}
The oracle replaces the perception module's object output with annotated semantic classes and 3D OBBs. Each box supplies a center, side lengths, and orientation axes in metric coordinates with positive-$z$ up. These boxes are the only ground-truth 3D representation supplied to the agent. Object counts are computed from the returned instances rather than read from a separate count annotation. The adapter supplies no question answers, question categories, ground-truth target-instance selections, dataset camera poses, or temporal annotations. A question identifier retrieves its scene; it does not select the objects or computation needed to answer that question. The agent retains responsibility for requesting classes, choosing instances and tools, and producing its answer.

\paragraph{Controlled inference and geometric adaptation.}
Both conditions use the same 32-frame policy, Qwen3-VL-8B-Instruct, system prompt, continuation instruction, model-visible tool schemas, decoding settings, and iteration limits. The oracle common-policy audit passed for all 4,030 selected questions, and all are retained in scoring. Two questions did not request the scene tool and remain included. The reasoning tool set is unchanged. Local-frame computation is unchanged.

\paragraph{Scope of the intervention.}
The oracle response lists the available annotated classes and explains the coordinate convention and annotation coverage. It cannot provide source-frame correspondences, appearance descriptions, or detection confidence, so it omits confidence-based default-selection fields. Room, navigation, and temporal operations return a uniform unsupported response while their schemas remain visible to every question. These differences mean the experiment diagnoses access to annotated object information, including identity and coverage, rather than changing only box coordinates. Annotation errors documented by ReVSI~\citep{zhang2026revsi} further mean that oracle accuracy measures compatibility with annotation-derived answers, not perfect recovery of physical geometry. The five evaluated categories and their scores are given in Table~\ref{tab:oracle}; room size, appearance order, and route planning are outside this diagnostic.

Both conditions evaluate the same 4,030 questions: 565 object-counting, 834 absolute-distance, 953 object-size, 710 relative-distance, and 968 relative-direction questions. Relative direction is averaged across its three official difficulty levels before computing the unweighted five-category average in Table~\ref{tab:oracle}. This diagnostic average is distinct from the eight-category average in Table~\ref{tab:vsi-bench}.

\clearpage
\section{Perception and Spatial Computation Details}
\label{sec:method_details}

\subsection{Metric Geometry and Co-refinement}
\label{sec:geometry_details}

Metric scale is estimated using robust depth-ratio statistics within frames, aggregated over reliable frames. Camera rotations transform the per-frame gravity predictions into reconstruction coordinates; cross-frame agreement removes inconsistent estimates before estimating a common up direction. If $\mathbf{u}_{\mathrm{raw}}$ is this direction, we choose $R$ satisfying $R\mathbf{u}_{\mathrm{raw}}=\mathbf{e}_z$ and transform points by $sR$. Camera geometry receives the same transformation, and the resulting scene uses $\mathbf{u}=\mathbf{e}_z$.

For same-class masks $M_i$ and $M_j$ in one frame, duplicate consolidation uses
\begin{equation}
\rho(M_i,M_j)=\frac{|M_i\cap M_j|}{\min(|M_i|,|M_j|)}>0.5.
\end{equation}
Masks satisfying this rule are united by pixelwise OR. For disjoint masks that may be parts of an occluded object, we retrieve same-class candidates using overlap between their gravity-aligned horizontal voxel projections. The judge then sees the source image with the candidate parts marked A/B and considers their structure, occlusion, appearance, and surrounding context. A positive judgment combines the masks and records the number of confirmed visible parts. Duplicate masks covering the same part do not increase this count.

After mask consolidation, the geometry-validity mask combines reconstruction confidence, depth-discontinuity checks, and global point-outlier filtering. Intersecting it with the consolidated mask yields the merged proposal's point candidates. Component selection is applied to this merged point set and retains the $k$ largest connected 3D components for $k$ confirmed parts, or all available components if fewer than $k$ remain. Ordinary masks use $k=1$. This choice preserves validated disconnected parts; it cannot distinguish arbitrary foreground and background components of equal size, nor recover points rejected by preceding filters.

\subsection{Geometry Filtering}
\label{sec:geometry_filtering}

The geometric validity set $V_t$ in Equation~\ref{eq:filtered_proposal} records which pixels in frame $t$ pass reconstruction-confidence, depth-edge, and 3D outlier filtering. Let $\Omega_t$ be the image grid, $D_t(\mathbf{x})$ the reconstructed depth in camera coordinates, and $\mathbf{p}_t(\mathbf{x})$ the corresponding point in the common reconstruction coordinates. The thresholds below are the defaults of the Pi3-based perception implementation.

\paragraph{Reconstruction-confidence filtering.}
For the confidence logit $\ell_t(\mathbf{x})$ predicted by Pi3, we retain pixels whose sigmoid confidence exceeds $\tau_c$:
\begin{equation}
V_t^{\mathrm{conf}}=
\left\{\mathbf{x}\in\Omega_t:\sigma\!\left(\ell_t(\mathbf{x})\right)>\tau_c\right\},
\qquad \tau_c=0.08.
\label{eq:confidence_filter}
\end{equation}
This confidence describes the reconstructed geometry; segmentation-mask confidence is not used in this test.

\paragraph{Depth-edge filtering.}
Let $\mathcal{N}_3(\mathbf{x})$ denote the $3\times3$ neighborhood centered at $\mathbf{x}$, restricted to the image grid. We compute the local depth range relative to the center pixel's depth and reject pixels exceeding $\tau_d$:
\begin{align}
e_t(\mathbf{x})&=
\frac{\max_{\mathbf{y}\in\mathcal{N}_3(\mathbf{x})}D_t(\mathbf{y})
-\min_{\mathbf{y}\in\mathcal{N}_3(\mathbf{x})}D_t(\mathbf{y})}
{D_t(\mathbf{x})},\nonumber\\
V_t^{\mathrm{depth}}&=
\left\{\mathbf{x}\in\Omega_t:e_t(\mathbf{x})\leq\tau_d\right\},
\qquad \tau_d=0.02.
\label{eq:depth_edge_filter}
\end{align}
This removes pixels near sharp depth transitions, where foreground and background projections can mix. The depth is taken from the reconstruction model's camera-coordinate point map.

\paragraph{Global 3D outlier filtering.}
We pool the reconstructed points passing both preceding filters across all frames into $\mathcal{Q}=\{\mathbf{q}_i\}_{i=1}^{L}$, retaining one sample per surviving pixel. From this pool, we estimate the empirical mean $\boldsymbol\mu$ and covariance $\Sigma$, and calculate squared Mahalanobis distances:
\begin{align}
a_i&=(\mathbf{q}_i-\boldsymbol\mu)^{\top}\Sigma^{\dagger}(\mathbf{q}_i-\boldsymbol\mu),\nonumber\\
\bar a&=\frac{1}{L}\sum_{i=1}^{L}a_i,
\qquad s_a=\sqrt{\frac{1}{L}\sum_{i=1}^{L}(a_i-\bar a)^2}.
\label{eq:global_outlier_filter}
\end{align}
Here $\Sigma^{\dagger}$ denotes the covariance pseudoinverse, equal to its inverse when nonsingular. We retain points satisfying $a_i<\bar a+\lambda s_a$, with $\lambda=5$. Thus, with $a(\mathbf{p})=(\mathbf{p}-\boldsymbol\mu)^{\top}\Sigma^{\dagger}(\mathbf{p}-\boldsymbol\mu)$, the final valid-pixel set is
\begin{equation}
V_t=\left\{\mathbf{x}\in V_t^{\mathrm{conf}}\cap V_t^{\mathrm{depth}}:
 a\!\left(\mathbf{p}_t(\mathbf{x})\right)<\bar a+5s_a\right\}.
\label{eq:geometry_validity_set}
\end{equation}
The mean, covariance, and outlier threshold are shared across the scene rather than estimated separately for each object. Intersecting $V_t$ with the consolidated instance mask $M_{t,i}$ selects that instance's geometrically filtered points. A coherent patch of background points can still pass these tests; the subsequent instance-level component selection addresses such residual contamination.

\subsection{Cross-view Instance Fusion}
\label{sec:association_details}

\paragraph{Geometric association settings.}
For the overlap score in Equation~\ref{eq:voxel_overlap}, the base configuration uses 0.1\,m voxels and $\tau_{\mathrm{ov}}=0.1$. Union--find maintains each group's source-frame set to implement Equation~\ref{eq:same_frame_exclusion}; an accepted merge updates that set by union. Small-instance filtering follows geometric fusion and precedes visual fragment association; the latter does not restore discarded instances.

\paragraph{Fragment candidate retrieval.}
For two sampled point sets, we calculate each point's nearest-neighbor distance to the other set in both directions. We use the smaller of the two directional fifth-percentile distances as a robust surface-proximity measure. The same computation on the gravity-aligned horizontal projections provides a second retrieval channel. Each fragment contributes its top-three same-class neighbors from each channel, and their union defines the candidate pairs.

\paragraph{Judge inputs and group verification.}
Each candidate pair is represented by up to three masked source views per instance, together with oblique and top-down renderings of the pair. Available source-camera poses are included in the 3D visualization. The fragments remain fixed in their shared coordinate frame: neither is independently translated, rotated, or rescaled to improve their apparent structural fit. Mask and point-cloud colors identify the candidates and are not evidence of their true appearance. The judge assesses \emph{3D structure}---compatible position, orientation, and structural arrangement---and \emph{appearance}---materials, textures, and corresponding visible details---and returns \texttt{same}, \texttt{different}, or \texttt{unclear}.

Group verification requires explicit \texttt{same} decisions for every pair across two groups before joining them. Thus A--B and B--C positive decisions do not establish A--C automatically; a missing, negative, or uncertain judgment blocks that group merge. The source-frame constraint in Equation~\ref{eq:same_frame_exclusion} continues to apply.

\subsection{Reasoning Tools}
\label{sec:reasoning_details}

The tool suite counts instances, measures object and room dimensions, computes inter-object distances from OBB geometry, determines spatial relations, plans routes, and locates object appearance times using source observations. The VLM selects objects and operations through structured tool calls. If a query requires object facing or an observer pose, these attributes are estimated through perception and supplied to the corresponding computation. View-dependent queries use the explicit observer frames described below.

\paragraph{Observer reference frames.}
\label{sec:reference_frame_details}

For a horizontal directional query, an observer at $\mathbf{a}$ faces an object at $\mathbf{b}$, and the unit scene-up vector is $\mathbf{u}$. We compute
\begin{align}
\mathbf{f} & = \operatorname{normalize}\!\left[(\mathbf{b}-\mathbf{a})
-\mathbf{u}\,\mathbf{u}^{\top}(\mathbf{b}-\mathbf{a})\right], &
\mathbf{r} & = \operatorname{normalize}(\mathbf{f}\times\mathbf{u}).
\end{align}
The observer basis is $B=[\mathbf{r},\mathbf{f},\mathbf{u}]$. For a target at $\mathbf{c}$, its observer-frame coordinates are $\mathbf{q}=B^{\top}(\mathbf{c}-\mathbf{a})$; their signs indicate right/left, front/back, and above/below.

For full 3D scene simulation, we retain the complete viewing direction: $\mathbf{f}=\operatorname{normalize}(\mathbf{b}-\mathbf{a})$. We compute $\mathbf{r}=\operatorname{normalize}(\mathbf{f}\times\mathbf{u})$ and use the orthogonalized up axis $\mathbf{u}'=\mathbf{r}\times\mathbf{f}$, yielding $B=[\mathbf{r},\mathbf{f},\mathbf{u}']$. A zero horizontal viewing direction is invalid for horizontal queries. For full 3D queries, coincident observer and facing locations are invalid. When the viewing direction is parallel to scene up, a fixed fallback reference axis determines the otherwise unspecified roll, and use of this convention is recorded.

\section{Baseline Model References}
\label{sec:baseline_models}

\subsection{Model References for Table~\ref{tab:mindcube-viewspatial}}
\label{sec:baseline_models_table3}

\paragraph{Base models.}
The direct VLMs and backbones appearing in Table~\ref{tab:mindcube-viewspatial} are Qwen2.5-VL-3B and Qwen2.5-VL-7B~\citep{bai2025qwen25vl}, Qwen3-VL-8B and Qwen3-VL-Thinking~\citep{bai2025qwen3vl}, GPT-4o~\citep{openai2024gpt4o}, GPT-5.4~\citep{openai2026gpt54}, Gemini-2.5-Pro~\citep{geminiteam2025gemini25}, Gemini-3-Pro~\citep{google2025gemini3}, InternVL2.5-8B~\citep{chen2024internvl25}, and InternVL3-8B~\citep{zhu2025internvl3}.

\paragraph{MindCube-Tiny methods.}
The methods in Table~\ref{tab:mindcube-viewspatial}(a) include VADAR~\citep{marsili2025visual} and pySpatial~\citep{luo2026pyspatial} with GPT-4o, GCA~\citep{chen2025gca} with Qwen3-VL-Thinking, AlloSpatial~\citep{ruan2026allospatial} with Gemini-3-Pro, and Think3D~\citep{zhang2026think3d} with Gemini-2.5-Pro. The VADAR citation identifies the original method; its MindCube-Tiny results are taken from the reproduction reported by \citet{luo2026pyspatial}.

\paragraph{ViewSpatial methods.}
The methods in Table~\ref{tab:mindcube-viewspatial}(b) include SenseNova-SI~\citep{cai2025sensenova} with InternVL3-8B, S-Agent~\citep{dai2026sagent} with GPT-5.4, MV-STRIDE~\citep{xu2026mvstride} with Qwen3-VL-8B, and CaMo~\citep{huang2026camo} and SpatialLadder~\citep{li2025spatialladder} with Qwen2.5-VL-3B. DeSpatial is the method presented in this paper and uses Qwen3-VL-8B in both subtables.

\subsection{Model References for Table~\ref{tab:vsi-super}}
\label{sec:baseline_models_table4}

Table~\ref{tab:vsi-super} compares DeSpatial with MovieChat~\citep{song2024moviechat}, Flash-VStream~\citep{zhang2024flashvstream}, Cambrian-S-7B and Cambrian-S-7B-LFP~\citep{yang2025cambrian}, and S-Agent~\citep{dai2026sagent}. The published baseline scores are taken from \citet{dai2026sagent}.

\section{Evaluation of Gravity Estimation}
\label{sec:gravity_human_evaluation}

\subsection{Human Assessment}

\begin{figure}[htbp]
\centering
\includegraphics[width=\linewidth]{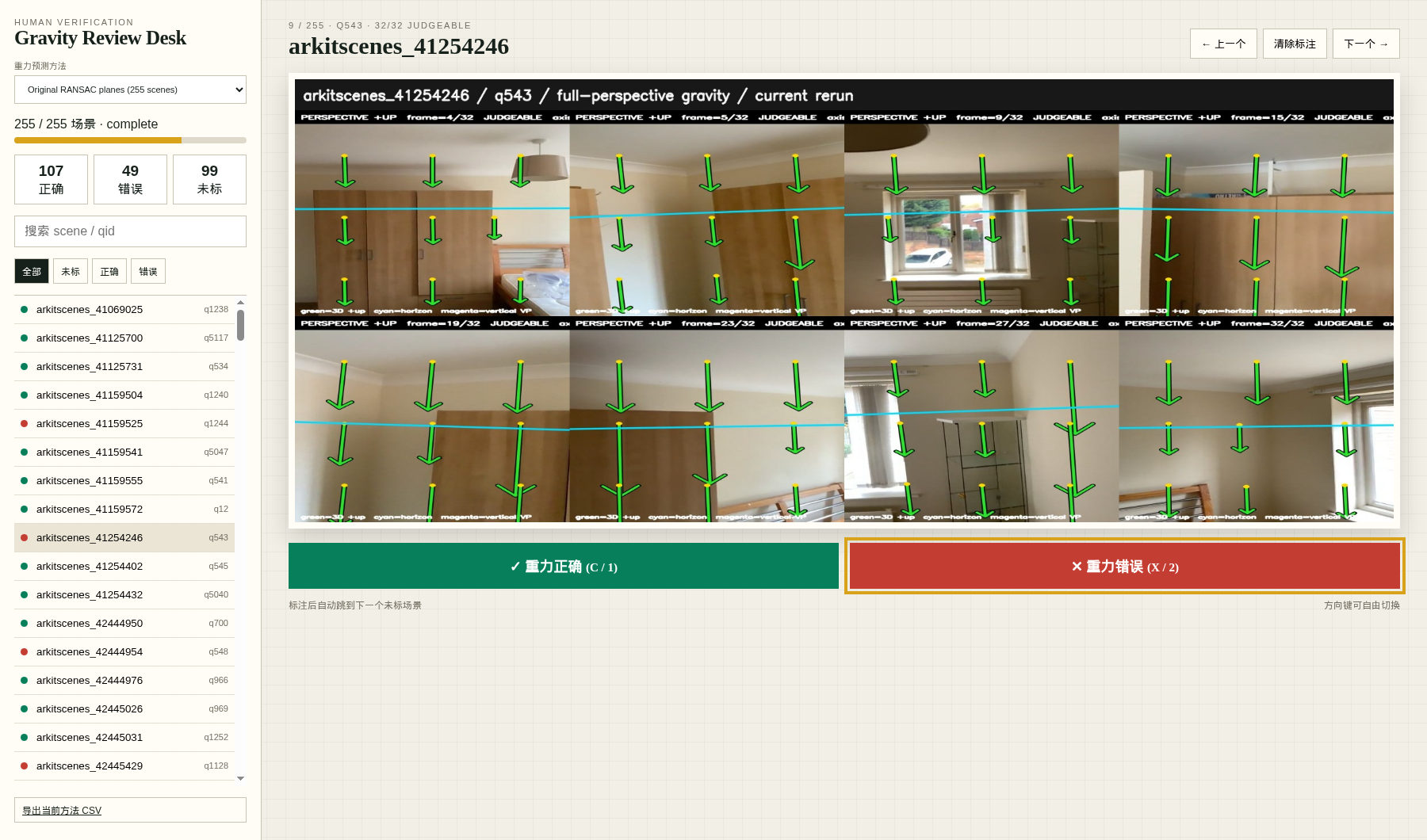}
\par{\small (a) RANSAC plane-based estimator: labeled incorrect.}\par
\includegraphics[width=\linewidth]{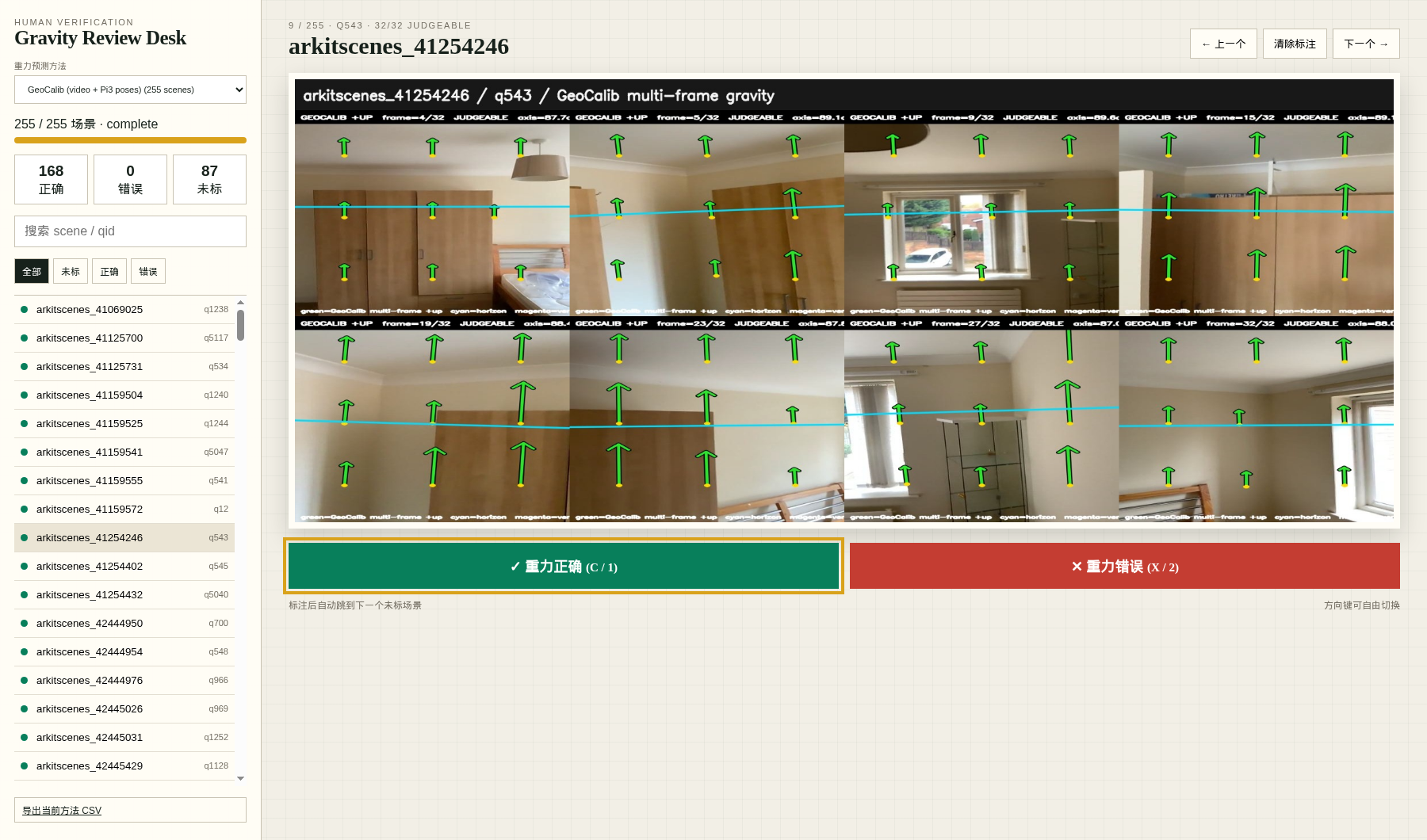}
\par{\small (b) GeoCalib joint calibration: labeled correct.}\par
\caption{Human annotation interface for gravity estimation. Both panels show the same scene and eight source views. Green arrows visualize the estimated up direction, and the highlighted button indicates the saved scene-level judgment.}
\label{fig:gravity_human_review}
\end{figure}

For each scene and method, we project the estimated scene-up direction onto multiple RGB frames using the corresponding camera intrinsics and poses, and display the projections as green arrows. The interface supports scene browsing and switching between the two named methods (Figure~\ref{fig:gravity_human_review}). The reviewer inspects the arrow orientations against the visible scene orientation across the displayed views and assigns one binary label, \emph{correct} or \emph{incorrect}, to that scene for the selected method. The green and red buttons record these two judgments, respectively.

Table~\ref{tab:gravity_ablation} reports the fraction of scenes labeled correct over the 156 scenes with annotations for both methods. This is a scene-level visual assessment of the projected up directions. The interface counters refer to the full 255-scene review pool, whereas the reported comparison uses only the paired 156-scene set. Figure~\ref{fig:gravity_human_review} shows the same eight frames of one scene for both methods, with the recorded labels \emph{incorrect} for RANSAC and \emph{correct} for GeoCalib.

\subsection{Historical Relative-Direction Diagnostic}
\label{sec:gravity_direction_diagnostic}

To examine the downstream relevance of gravity estimation, we revisit two runs before and after introducing GeoCalib: both use Qwen3-VL-8B-Instruct and 32 frames per video on the same 267 relative-direction questions from 105 VSI-Bench videos; question IDs and associated videos match exactly. Table~\ref{tab:gravity_direction_history} reports the archived scores. The relative-direction score increases from 56.17 to 71.75, a gain of 15.58 points. In the paired answers, 55 questions change from incorrect to correct and 13 change from correct to incorrect.

\begin{table}[htbp]
\centering
\small
\setlength{\tabcolsep}{5pt}
\begin{tabular}{lrrrrr}
\toprule
Development run & Easy & Medium & Hard & Rel. Dir. & Correct / Total \\
\midrule
RANSAC-based (Jul. 11) & 63.70 & 48.72 & 56.10 & 56.17 & 158 / 267 \\
GeoCalib-based (Jul. 20) & 81.51 & 66.67 & 67.07 & 71.75 & 200 / 267 \\
\bottomrule
\end{tabular}
\caption{Historical relative-direction scores (\%) on identical questions. Easy, Medium, and Hard contain 146, 39, and 82 questions, respectively; Rel. Dir. is their unweighted mean. This development comparison does not isolate gravity as the sole intervention.}
\label{tab:gravity_direction_history}
\end{table}

\paragraph{Up-sign diagnostic.}
An initial analysis of the earlier run identified 66 incorrectly answered questions for which reversing the estimated up direction, while keeping the selected objects and their coordinates fixed, makes the tool-computed direction agree with the ground-truth answer. This observation motivated us to inspect the estimated gravity directions more closely: in the horizontal observer frame of Appendix~\ref{sec:reference_frame_details}, reversing the scene-up direction reverses the right axis while preserving the forward axis, exchanging left and right. After introducing GeoCalib, 45 of these 66 questions are answered correctly. Thus, 45 of the 55 improved answers (81.8\%) coincide with cases identified by this up-sign diagnostic.